\documentclass[lettersize,journal]{IEEEtran}

\usepackage{amsmath,amsfonts}
\usepackage{array}
\usepackage{caption}
\usepackage{textcomp}
\usepackage{stfloats}
\usepackage{url}
\usepackage{array}
\usepackage{makecell}
\usepackage{multirow}
\usepackage{verbatim}
\usepackage{graphicx}
\usepackage{cite}
\usepackage{hyperref}
\usepackage{bm}
\usepackage{xfrac}
\usepackage{mathtools}
\usepackage{enumitem}
\usepackage{titlesec}
\usepackage{subcaption}
\usepackage[ruled, lined, linesnumbered, commentsnumbered, longend]{algorithm2e}
\hypersetup{colorlinks=true,
            linkcolor=magenta,
            urlcolor=magenta,
            citecolor=green}
\newcommand{\PreserveBackslash}[1]{\let\temp=\\#1\let\\=\temp}
\newcolumntype{C}[1]{>{\PreserveBackslash\centering}p{#1}}
\newcolumntype{R}[1]{>{\PreserveBackslash\raggedleft}p{#1}}
\newcolumntype{L}[1]{>{\PreserveBackslash\raggedright}p{#1}}

\begin{document}

\title{One Step Diffusion-based Super-Resolution with Time-Aware Distillation}

\author{Xiao He$^{1}$, Huaao Tang$^{2}\thanks{$^\dagger$Corresponding Author.}$, Zhijun Tu$^{2}$, Junchao Zhang$^{2}$, Kun Cheng$^{1}$, Hanting Chen$^{2}$, Yong Guo$^{3}$, Mingrui Zhu$^{1}$, Nannan Wang$^{1\dagger}$, Xinbo Gao$^{4}$, Jie Hu$^{2}$ \\
\small$^1$ State Key Laboratory of Integrated Services Networks, School of Telecommunications Engineering, Xidian University. \\
  \small$^2$ Huawei Noah's Ark Lab. \\
  \small$^3$ Consumer Business Group, Huawei. \\
  \small$^4$  Chongqing Key Laboratory of Image Cognition, Chongqing University of Posts and Telecommunications. \\
  \small\texttt{nnwang@xidian.edu.cn}\\
}
\markboth{}%
{Shell \MakeLowercase{\textit{et al.}}: A Sample Article Using IEEEtran.cls for IEEE Journals}

\maketitle

\begin{abstract}
Diffusion-based image super-resolution (SR) methods have shown promise in reconstructing high-resolution images with fine details from low-resolution counterparts. However, these approaches typically require tens or even hundreds of iterative samplings, resulting in significant latency. Recently, techniques have been devised to enhance the sampling efficiency of diffusion-based SR models via knowledge distillation. Nonetheless, when aligning the knowledge of student and teacher models, these solutions either solely rely on pixel-level loss constraints or neglect the fact that diffusion models prioritize varying levels of information at different time steps. To accomplish effective and efficient image super-resolution, we propose a time-aware diffusion distillation method, named TAD-SR. Specifically, we introduce a novel score distillation strategy to align the data distribution between the outputs of the student and teacher models after minor noise perturbation. This distillation strategy enables the student network to concentrate more on the high-frequency details. Furthermore, to mitigate performance limitations stemming from distillation, we integrate a latent adversarial loss and devise a time-aware discriminator that leverages diffusion priors to effectively distinguish between real images and generated images. Extensive experiments conducted on synthetic and real-world datasets demonstrate that the proposed method achieves comparable or even superior performance compared to both previous state-of-the-art (SOTA) methods and the teacher model in just one sampling step. Codes are available at \url{https://github.com/LearningHx/TAD-SR}. 
\end{abstract}


\section{Introduction}
\IEEEPARstart{I}mage super-resolution (SR), a cornerstone task in low-level vision, involves reconstructing high-resolution (HR) images with intricate details from low-resolution (LR) counterparts. Owing to the inherent ill-posed nature of this task, as multiple high-resolution reconstructions are plausible for a given low-resolution input, presenting a persistent and perplexing challenge. Recently, the diffusion model \cite{ho2020denoising,song2020denoising}, a novel generative model, has garnered increasing attention for its capacity to model complex data distributions. It has gradually emerged as a successor to Generative Adversarial Networks (GANs) \cite{goodfellow2020generative} in various downstream tasks, including image editing \cite{meng2021sdedit,hertz2022prompt}, image inpainting \cite{chung2022come,lugmayr2022repaint} and image super-resolution \cite{saharia2022image,yue2024resshift}.

Specifically, existing diffusion-based image super-resolution methods can be broadly categorized into two streams: one involves feeding low-resolution images along with noise into the diffusion model as input \cite{saharia2022image,yue2024resshift}, while the other \cite{choi2021ilvr,fei2023generative} adapts SR tasks by modifying the sampling process on a pre-trained diffusion model. While these methods have demonstrated promising results, generating HR images typically demands tens or even hundreds of iterative samplings, significantly impeding their practical application and further advancement.

\begin{figure*}
\centerline{\includegraphics[width=1.0\linewidth]{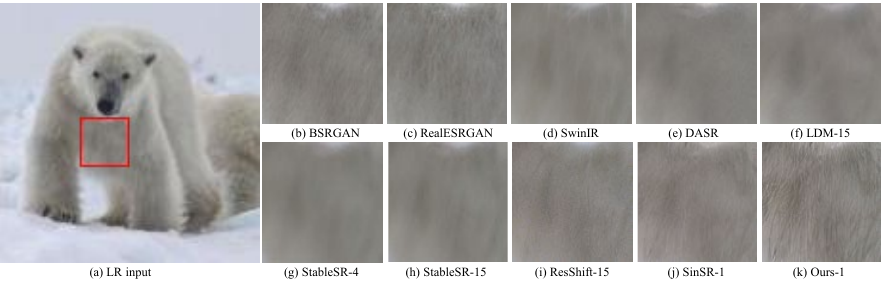}}
\label{fig:intro}
\caption{ Qualitative comparisons on one typical real-world example of the proposed method and recent state of the arts, including BSRGAN \cite{zhang2021designing}, RealESRGAN \cite{wang2021real}, SwinIR \cite{liang2021swinir}, DASR \cite{liang2022efficient}, LDM \cite{rombach2022high}, StableSR \cite{wang2023exploiting}, ResShift \cite{yue2024resshift}, and SinSR \cite{wang2023sinsr}. We mark the number of sampling steps of diffusion-based SR method with the format of “ Method-A” for more intuitive visualization, where “A” is the number of sampling steps. Note that LDM contains 1000 diffusion steps in training and is accelerated to “A” steps using DDIM \cite{song2020denoising} during inference. Please zoom in for a better view.}
\vspace{-2mm}
\end{figure*}

To enhance the inference efficiency of diffusion models, various acceleration techniques have been proposed, such as the development of numerical samplers \cite{lu2022dpm,zheng2024dpm} and the applications of knowledge distillation \cite{salimans2022progressive,sauer2023adversarial}. However, employing numerical solvers to accelerate sampling speed often compromises performance. Moreover, the majority of methods employing knowledge distillation are tailored for text-to-image generation, where ensuring the diversity of generated images is crucial. In contrast, super-resolution tasks necessitate output images with high visual fidelity relative to LR images. Therefore, directly applying these methods to super-resolution tasks presents significant challenges. For the SR task, ResShift \cite{yue2024resshift} has improved the sampling efficiency of diffusion-based SR models by utilizing information from LR images to reformulate the diffusion process, thereby reducing the number of sampling steps to 15. Furthermore, SinSR \cite{wang2023sinsr} merges distillation techniques with a cycle consistency approach to refine the ResShift model into a single inference step. Nonetheless, it still yields blurry results in the high-frequency details of the image, despite ensuring the visual fidelity of the generated image through pixel-level loss constraints. Recently, AddSR \cite{xie2024addsr} employs adversarial diffusion distillation (ADD) \cite{sauer2023adversarial} for SR task to enhance sampling efficiency while ensuring the perceptual quality of generated images. However, it inherits the issue of image oversaturation present in score distillation sampling. Additionally, it relies on a pre-trained DINOV2 discriminator in pixel space, which is both expensive and challenging to optimize. 

To address the issues in the aforementioned methods, we introduce a novel time-aware distillation method to accelerate the sampling rate of diffusion-based SR models. It enables the generation of high-resolution images with fine details in a single sampling step. Specifically, considering that diffusion models focus on high-frequency information at small time steps, we have designed a high-frequency enhanced score distillation strategy. This strategy refines the student network by evaluating the difference in the distribution of the teacher model and student model outputs after mild noise disturbance, thereby improving the high-frequency details in the generated images of the student model. To overcome the performance limitations of teacher models, we incorporate generative adversarial learning into the distillation framework, forcing the student model to directly generate samples that lie on the manifold of real images in a single inference step. Specifically, we engineer a time-aware discriminator within the latent space that is capable of differentiating between the data distributions of authentic and synthetically generated images subjected to various perturbations. Experimental evidence confirms that the introduction of temporal information into the discriminator significantly enhances its ability to more effectively regulate the generator's performance.

Overall, our contributions can be summarized as follows:
\begin{itemize}[topsep=0pt,parsep=0pt,leftmargin=18pt]
  \item We introduce a novel time-aware distillation method that accelerates the diffusion-based SR model into a single inference step.
  \item To improve the high-freauency details of the generated image, we introduce a novel score distillation method that refines the student model by quantifying the discrepancy in scores between the outputs of the teacher and student models at small time steps. 
  \item Furthermore, to ensure the performance of student model is not solely bound by the teacher model, we incorporate generative adversarial learning into the distillation framework. A key innovation in our approach is the development of a time-aware discriminator. This discriminator is capable of distinguishing between the data distributions of real and generated images that have undergone various perturbations in latent space.
  \item Extensive experiments on both general and facial datasets have demonstrated that our method, using only single-step sampling, achieves performance that is comparable to or surpasses state-of-the-art methods and teacher models.

\end{itemize}

\section{Related Work} \label{sec:related-work}
\subsection{Image Super-Resolution.} Traditional methods \cite{dong2012nonlocally,gu2017weighted,gu2015convolutional} for image super-resolution rely on manual design of image priors based on subjective knowledge to restore image details. With the advancement of deep learning (DL), DL-based image super-resolution has become predominant, which mainly focused on network architecture \cite{lai2017deep,menick2018generating,lugmayr2020srflow,sajjadi2017enhancenet}, image priors \cite{pan2021exploiting,chan2021glean}, loss functions \cite{zhou2020guided,fuoli2021fourier}, and other aspects \cite{zhang2018learning,wang2021real}. Recently, diffusion-based methods for image super-resolution have garnered widespread attention. SR3 \cite{saharia2022image} incorporated low-resolution images as conditions into the denoising model to guide the sampling process. Subsequently, CDPMSR \cite{niu2023cdpmsr} and IDM \cite{gao2023implicit} respectively utilized preprocessed images and features as conditions to enhance the perceptual quality. In order to reduce the computational cost of the model, StableSR \cite{wang2023exploiting} and PASD \cite{yang2023pixel} utilized the powerful generation priors of stable diffusion (SD) \cite{rombach2022high} to achieve image super-resolution in latent space through diffusion processes. However, these methods typically require dozens or even hundreds of iterations to generate high-resolution images. To enhance the inference efficiency, ResShift \cite{yue2024resshift} redesigned the diffusion process by shifting the residuals between high-resolution and low-resolution images to construct a Markov chain, achieving performance comparable to previous state-of-the-art methods with just 15 sampling steps. 

\subsection{Accelerating Diffusion Models. }Although diffusion model \cite{ho2020denoising,rombach2022high} has formidable generation capabilities, the substantial number of inference steps poses a significant obstacle to its practical implementation leading to methods reducing the inference steps. Mainstream approaches include the development of high-order samplers \cite{song2020denoising,lu2022dpm,zheng2024dpm} and the application of knowledge distillation techniques \cite{salimans2022progressive,sauer2023adversarial,sauer2024fast,song2023consistency,luo2023latent}. Denoising diffusion implicit models (DDIM) \cite{song2020denoising}, an early contribution, introduced a deterministic sampling method that notably decreased the number of diffusion sampling steps. DPMSolver \cite{lu2022dpm} proposed a fast dedicated high-order ODE solver, further reducing the diffusion sampling steps to 20. However, trajectory compression through numerical solvers often results in performance degradation, necessitating over ten inference steps to generate samples. In contrast, progressive distillation \cite{salimans2022progressive} gradually reduces the inference steps of student models through multi-stage distillation, but the accumulation of errors in each distillation stage may affect the performance of the student model. Consistency model \cite{song2023consistency} eliminates the need for computation-intensive iterations by applying consistency regularization to ODE trajectories. Additionally, Adversarial diffusion distillation (ADD) \cite{sauer2023adversarial} integrates generative adversarial networks with score distillation to enhance the perceptual quality of student network-generated images. For image super-resolution tasks, AddSR \cite{xie2024addsr} introduces two key advancements based on adversarial distillation technology, effectively fulfilling image super-resolution objectives. Inspired by cycle consistency loss, SinSR \cite{wang2023sinsr} proposes a single-step image super-resolution method. However, AddSR overlooks the influence of time steps on the discriminator, while SinSR primarily focuses on constraining latent codes through pixel-level loss, neglecting perceptual distribution alignment. To achieve image super-resolution more efficiently and effectively, this work propose a time-aware diffusion distillation method.

\begin{figure*}
\centerline{\includegraphics[width=1.0\linewidth]{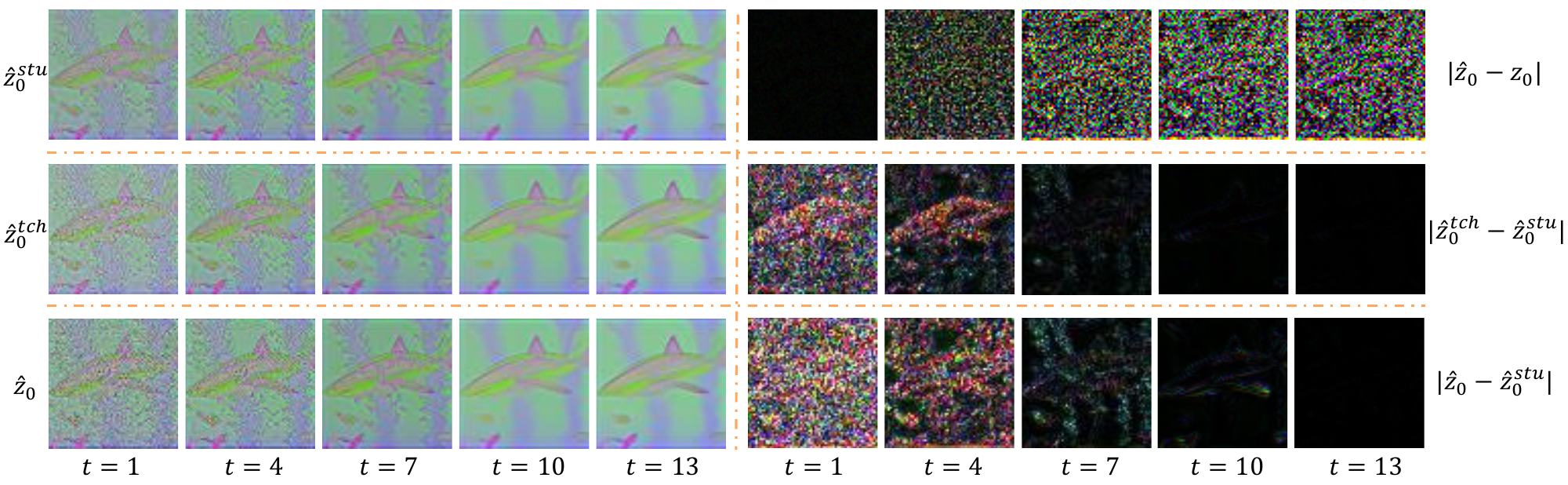}}
\caption{We inject varying degrees of noise into the outputs of the student model, the teacher model, and the corresponding real data (HR images). These noisy data are then fed into a pre-trained diffusion-based SR model to obtain their denoising scores(visualized as clean data prediction $\hat{z}_0$).}
\label{fig:motivation}
\end{figure*}

\section{Preliminary}
\textbf{Diffusion models} is a type of probabilistic generative model, which utilize a Markov chain transforms complex data distribution $z_0\sim p_{data}$ into noise distribution $z_T\sim \mathcal{N}\left(0, I \right)$ and recover the data by gradually removing the noise. In image supre-resolution tasks, Resshift \cite{yue2024resshift} changes the initial state of diffusion model and constructs a new Markov chains to generate high-resolution images. The forward process can be mathmatically expressed as follows:
\begin{equation}
    q\left(z_t| z_0,y\right) = \mathcal{N} \left(z_t \vert z_0 + \eta_t\left(z_y-z_0\right),\kappa ^2 \eta_t I\right)
    \label{eq:forward}
\end{equation}
where $z_0$ and $z_y$ represent the latent codes obtain by encoding the HR images $x$ and LR images $y$, respectively.
$\eta_t$ is a serial of hyper-parameters that monotonocally increases with timestep $t$ and satifies $\eta_0 \rightarrow 0$ and $\eta_T \rightarrow 1$, $\kappa$ is a hyper-parameter controlling the noise variance. Baed on this forward process, the reverse process will commence from the initial sate with rich information in low-resolution images to perform denoising. The formula is as follows:
\begin{equation}
  q\left(z_{t-1}| z_t,z_0,y\right) = \mathcal{N} \left(z_{t-1}\big\vert \frac{\eta_{t-1}}{\eta_t}z_t + \frac{\alpha_t}{\eta_t} z_0,\kappa ^2 \frac{\eta_{t-1}}{\eta_t} \alpha_t \bm{I}\right)
  \label{eq:backward}
\end{equation}
where $\alpha_t = \eta_{t}-\eta_{t-1}$, To mitigate the influence of randomness on distillation \cite{wang2023sinsr}, we reformulate Eq.~\ref{eq:backward} to employ deterministic sampling, as follows:

\begin{equation}
  q\left(z_{t-1}| z_t,z_0,y\right) = \delta \left(k_{t}z_{0} + m_{t}z_{t} + j_{t}z_y\right)
  \label{eq:backward_determine}
\end{equation}

where $\delta$ is the unit impulse, $m_t = \sqrt{\frac{\eta_{t-1}}{\eta_t}}$, $j_t = \eta_{t-1}-\sqrt{\eta_{t-1}\eta_{t}}$ and $k_t = 1-j_t-m_t$. The details of the derivation can be found in SinSR~\cite{wang2023sinsr}. In the backward process Eq.~\ref{eq:backward_determine}, $x_0$ is usually predicted by a trainable deep neural network $f_\theta$ with parameter $\theta$. The training objective function of $f_\theta$ as follows:

\begin{equation}
  \min_{\bm{\theta}} \sum\nolimits_t w_t \Vert f_{\bm{\theta}}(\bm{z}_t, \bm{y}, t) - \bm{z}_0 \Vert_2^2,
  \label{eq:diffusion_loss}
\end{equation}
where $w_t = \frac{\alpha_t}{2\kappa^2\eta_t\eta_{t-1}}$. In practice omitting this weight often leads to performance improvement~\cite{ho2020denoising}.

\textbf{Score Distillation Sampling (SDS)} is a distillation technique for pre-trained diffusion models, effectively applied in generating 3D assets \cite{poole2022dreamfusion,lin2023magic3d} and accelerating diffusion models \cite{sauer2023adversarial,sauer2024fast}. It leverages the rich generative prior of diffusion models to optimize the generated images or the geenrator, which can be expressed as follows:
\begin{equation}
	 \nabla_{\theta}\mathcal{L}_{SDS}(\bm{z},y,\epsilon,t) = (\epsilon_{\phi}(\bm{z}_t,y,t)-\epsilon)\frac{\partial{\bm{z}_t}}{\partial{\theta}}
   \label{eq:SDS}
\end{equation}
where $z_t$ refers to the noised version of $z$. According to \cite{poole2022dreamfusion}, the U-Net jacobian term $\frac{\partial{\epsilon_{\phi}(\bm{z},y,t)}}{\partial{\bm{z}_t}}$ is omitted in Eq.~\ref{eq:SDS} to lead an effective gradient.

\begin{figure*}
  \centerline{\includegraphics[width=1.0\linewidth]{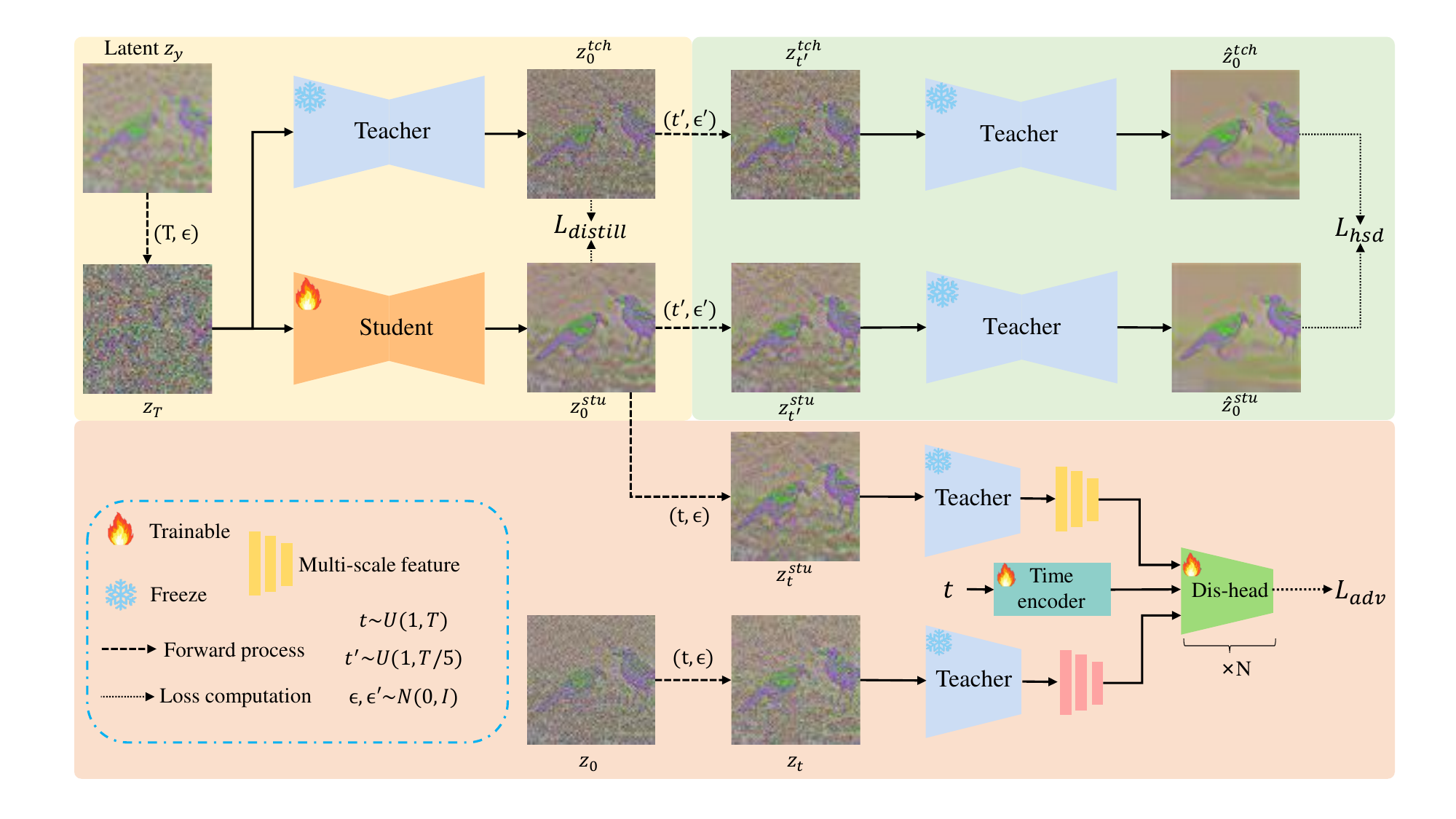}}
  \caption{\textbf{Method overview.} We train our student model to map a noisy latent $z_T$ to clean latent $z_0^{stu}$ through single step sampling. To match the student model's output with the multi-step sampling outputs of the teacher model, we optimize the student model using both vanilla distillation loss and our proposed high-frequency enhanced score distillation. Additionally, to further improve the performance of the student model, we propose a time-aware discriminator that provides effective supervision through generative adversarial training.}
  \label{fig:method}
  \end{figure*}

\section{Methodology}

\subsection{Motivation}
\label{subsec:motivation}

To explore distillation methods more applicable to accelerating diffusion-based SR models, we conducted an experiment to visualize the score differences among the single-step output of the student model, the multi-step output of the teacher model, and the high-resolution images following noise perturbation in latent space. (Note that the student model has the same weight as the teacher model.) Specifically, we re-noise the outputs of the student model, the teacher model, and the real data (HR image) to varying degrees, then feed the noise data into a pre-trained diffusion-based SR model to view the denoising scores (visualized as clean data prediction $\hat{z}_0$ in Fig.~\ref{fig:motivation}) and compare their differences of scores. 

From the first row of Fig.~\ref{fig:motivation}, it can be seen that even when real data are fed into a well-trained diffusion-based SR model, there is still a gap between the predicted scores $\hat{z}_0$ and the real scores $z_0$. This indicates that even in ideal situations, score-based distillation (SDS) itself has biases, consistent with the conclusions of previous related work \cite{hertz2023delta,wang2024prolificdreamer}. Therefore, utilizing SDS to accelerate diffusion-based SR models is not optimal. To identify the shortcomings of the student model and design a suitable distillation strategy, we further analyzed the score differences $|\hat{z}_0^{tch}-\hat{z}_0^{stu}|$ between the outputs of the teacher and student models after noise disturbance, as illustrated in the second row of Fig.~\ref{fig:motivation}. From the figure, it is evident that there is a notable difference in denoising scores between the output of the teacher model and that of the student model when subjected to minor noise disturbances. Additionally, diffusion models primarily focus on high-frequency information in images at small time steps. Based on these observations, it can be concluded that the student model's output mainly lacks high-frequency details compared to the teacher model. Therefore, we believe that computing the score differences between the outputs of the teacher and student models under mild noise disturbances can provide an effective gradient direction to guide the optimization of the student model.

In addition, the third row in Fig.~\ref{fig:motivation} visualizes the score difference $|\hat{z}_0-\hat{z}_0^{stu}|$ between the output of the student model and the real data after noise disturbance. It can be seen that this score difference is significant, and this difference changes with time steps. However, existing methods do not consider this factor when distinguishing the distribution of real data and generated data after noise disturbance. Instead, they directly use a discriminator to distinguish the differences between the data distributions of real and generated images subjected to various perturbations, which poses a challenge to the optimization of the discriminator.

\subsection{TAD-SR}
\label{sub:TAD-SR}
The overview framework of our proposed TAD-SR is illustrated in Fig.~\ref{fig:method}, consisting of a teacher model $F_{\phi}$ parameterized by $\phi$, a student network $f_{\theta}$ initialized from the teacher model with weights $\theta$, and a trainable time-aware discriminator $D_{\psi}$ parameterized by $\psi$. During training, the student model generates samples from noisy data and computes the mean squared loss against the samples generated iteratively by the teacher model. Subsequently, we introduce slight noise to the samples produced by both the student and teacher models, predict the score function via the teacher model, and refine the student network by leveraging the discrepancy between the two score functions. Furthermore, to mitigate the performance constraints of the teacher model on the student model, we design a time-aware discriminator built upon the encoder of the pre-trained teacher model, enhancing the perceptual quality of the generated samples through adversarial training processes. 

\textbf{Vanilla distillation.} We utilize the multi-step output results of the teacher model as the learning objective for the student model. It guides the student model to establish a mapping between low-resolution and high-resolution images through single-step inference. The distillation loss is formulated as follows:
\begin{equation}
  \mathcal{L}_{distill} = \| F_{\phi}\left(z_T,T,y\right) - f_{\theta}\left(z_T,T,y\right)\Vert_2^2
  \label{eq:distill}
\end{equation}
where $z_T$ is obtained through the forward process Eq.~(\ref{eq:forward}). Specifically, when time step $t=T$, noisy latent $z_T \sim  \mathcal{N} \left(x_t;y,\kappa ^2 \eta_t \bm{I}\right)$.

\begin{figure*}[htbp]
  \centering
  \begin{minipage}{0.32\textwidth}
      \centering
      \includegraphics[scale=0.6]{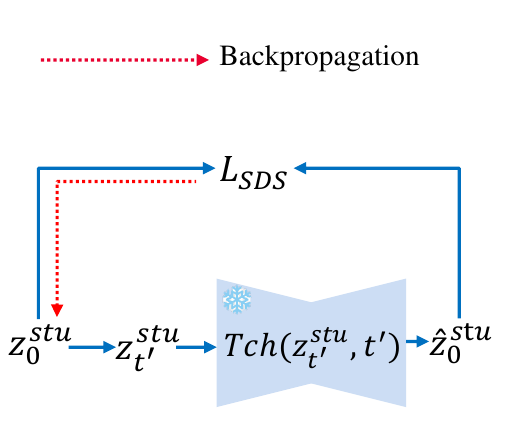}
      \subcaption{SDS \cite{poole2022dreamfusion}}
      \label{fig:subfig1}
  \end{minipage}\hfill
  \begin{minipage}{0.32\textwidth}
      \centering
      \includegraphics[scale=0.6]{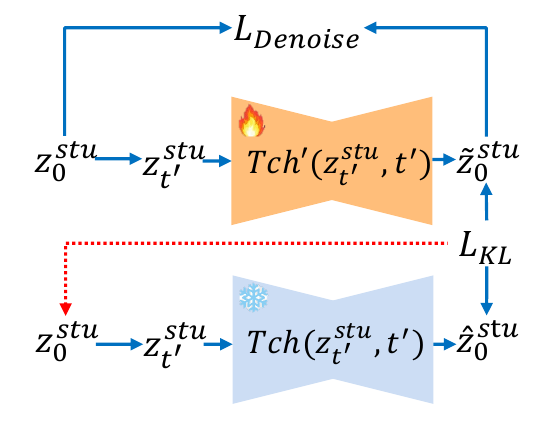}
      \subcaption{DMD \cite{yin2023one}}
      \label{fig:subfig2}
  \end{minipage}\hfill
  \begin{minipage}{0.32\textwidth}
      \centering
      \includegraphics[scale=0.6]{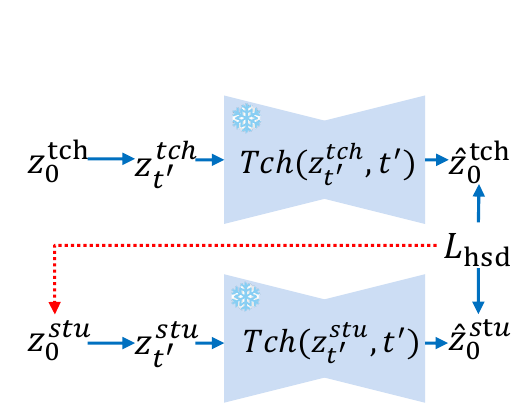}
      \subcaption{Ours}
      \label{fig:subfig3}
  \end{minipage}
  \caption{Comparison of various score distillation techniques. Compared to SDS \cite{poole2022dreamfusion,sauer2023adversarial} and DMD \cite{yin2023one}, our HF-enhanced score distillation fully utilizes the potential of teacher model, providing meaningful gradient guidance to student models without the need to retrain a new diffusion model.}
  \label{fig: TSD}
\end{figure*}

\textbf{High-frequency enhanced score distillation.} As analyzed in Section \ref{subsec:motivation}, employing SDS \cite{poole2022dreamfusion} to accelerate diffusion-based SR models is not an optimal solution. Its inherent bias may introduce meaningless gradient directions to the student model, leading to output blurring, oversaturation, and other issues \cite{wang2024prolificdreamer,hertz2023delta}. To eliminate this bias, DMD \cite{yin2023one} trains a new diffusion model to learn the score function of samples generated by the student model and updates the generator based on the difference between the score functions predicted by the new model and the teacher model. However, this approach involves a complex training process that requires alternating training between the student model and the new diffusion model. 

By contrast, based on the observations presented in Fig.~\ref{fig:motivation}, we develop an effective and efficient score distillation method. Specifically, we analyze the score differences between the outputs of the teacher model and the student model under various noise disturbances. Experiments indicate that these differences are primarily significant under mild noise disturbances (i.e., small time steps). Given that diffusion models typically predict high-frequency information in images at small time steps, this suggests that images generated by student models are predominantly deficient in high-frequency details compared to those produced by teacher models. Consequently, we propose a high-frequency enhanced score distillation method. This approach optimizes the student model by computing the score differences between the outputs of the teacher model and the student model under slight noise disturbances, thereby enhancing the student model's ability to accurately capture and reproduce high-frequency details. Specifically, we apply slight Gaussian noise to perturb the outputs of both the student and teacher models and model their data distributions through the teacher model. Subsequently, we optimize the student model by calculating the differences between these distributions. According to score distillation, the following formula is derived:

\begin{equation}
   \mathcal{L}_{hsd} = \mathbb{E}_{z_{t^{'}}^{tch},z_{t^{'}}^{stu},y} \left[\omega\left( \epsilon_{\phi}\left(z_{t^{'}}^{stu},t,y\right)-\epsilon_{\phi}\left(z_{t^{'}}^{tch},t,y\right)\right)\right],  
  \label{eq:HSD_V1}
\end{equation}
where $\omega = 1/CS $ is a weighting function, $C$ is the number of channels and $S$ is the number of spatial pixels. $z_{t^{'}}^{tch}$ and $z_{t^{'}}^{stu}$ are the noise data obtained by adding noise to the outputs of the teacher model $z_{0}^{tch}$ and the output of student model $z_{0}^{stu}$, respectively, through Eq.~\ref{eq:forward}. Since we primarily supplement the high-frequency details of the generated images of the student model, time step $t^{'}$ in Eq.~\ref{eq:forward} are uniformly sampled in $[1,3]$. Note that during the loss backpropagation in Eq.~\ref{eq:HSD_V1}, similar to SDS~\cite{poole2022dreamfusion}, we omit the U-Net Jacobian matrix term. According to Eq.~\ref{eq:forward}, we can simplify Eq.~\ref{eq:HSD_V1} as the following formula:

\begin{equation}
   \mathcal{L}_{hsd} =  \omega_2 \left(z_{0}^{stu}-z_{0}^{tch}+F_\phi\left(z_{t^{'}}^{tch},t,y\right) - F_\phi\left(z_{t^{'}}^{stu},t,y\right)\right).  
  \label{eq:HSD_V2}
\end{equation}
where $\omega_2 = \frac{\omega \left(1-\eta_{t^{'}}\right)}{\sqrt{\eta_{t^{'}}}\kappa}$. The details of the derivation can be found in the appendix.

From the above equation, it can be seen that when the output of the student model is the same as that of the teacher model, the loss is zero, and there is no additional bias. Compared to SDS \cite{poole2022dreamfusion,sauer2023adversarial}, our proposed score distillation provides more meaningful gradient guidance for student models.

\textbf{Time-aware adversarial loss.} ADD \cite{sauer2023adversarial} has demonstrated that combining diffusion models with generative adversarial networks can significantly improve the perceptual quality of generated images. However, this approach relies on pre-training the DINOv2 discriminator in pixel space, which is both costly and complex. To reduce training costs and enhance model performance, LADD \cite{sauer2024fast} employed a pre-trained diffusion model for generative adversarial training in latent space. Although LADD reduces the training costs associated with adversarial diffusion distillation, it overlooks the significant correlation between the features extracted by the diffusion model and the time steps. By employing only a single discriminator, which lacks any temporal information, to differentiate between the distribution differences of real and synthetic data across various noise disturbances, LADD introduces a challenge to the optimization of the discriminator. To address this issue, we propose a time-aware discriminator, which is capable of distinguishing between the data distributions of real and generated images that have undergone various perturbations in latent space. Specifically, we first utilize the encoder part of the teacher model to extract multi-scale features $F_k$ from both the student model's output images and real images.

\begin{equation}
  F_k = Enc_{\phi} \left(z_t,t,y\right),
\end{equation}
where $Enc_{\phi}$ denotes the encoder part of the teacher model, $z_t$ represents the noisy latent code after adding noise to the student model output or the real latent code. We use $F_k^{stu}$ and $F_k^{gt}$ to denote the features extracted from real latent code and the features extracted from the output of the student model, respectively. We then encode the time step $t$ and map it to a set of time embeddings corresponding to the multi-scale feature dimensions using several linear layers. These time embeddings are used to modulate the features:

\begin{equation}
  \hat{F}_{k}= Norm\left(F_k\right) * (1+\gamma_k)+ \beta_k,
\end{equation}
where $\gamma_k$ and $\beta_k$ is a set of learnable modulation parameters produced by projecting time step $t$ into embedding space. After modulation, features of different scales are sent to various discrimination heads $D_{\psi,k}$ for evaluation. The final output is obtained by averaging the results from each discrimination head. The corresponding adversarial loss can be formulated as follows:

\begin{equation}
\mathcal{L}_{adv}^{f_{\theta}} = -\mathbb{E}_{z_0^{stu}} \left[\sum_{k} D_{\psi,k}\left(\hat{F}_{k}^{stu}\right) \right],
\label{eq:adv_G}
\end{equation}

\begin{align}
\mathcal{L}_\text{adv}^{D_{\psi}} &=\mathbb{E}_{z_0^{stu}} \left[\sum_{k}\text{max}\left(0, 1+D_{\psi,k}\left(\hat{F}_{k}^{stu}\right)\right) \right] \notag \\ 
&+ \mathbb{E}_{z_{0}} \left[\sum_{k}\text{max}\left(0, 1-D_{\psi,k}\left(\hat{F}_{k}\right)\right) \right]
\label{eq:adv_D}
\end{align}

\textbf{The total objective.} The student network is trained with the above three losses as follows:
\begin{equation}
  \mathcal{L}_{f_\theta}  = \mathcal{L}_{distill} + \lambda_1 \mathcal{L}_{hsd} + \lambda_2 \mathcal{L}_{adv}^{f_{\theta}},
\end{equation}
where $\lambda_1$ and $\lambda_2$ are the hyperparameters to control the relative importance of these objectives.

\begin{table*}[t]
  \centering
  \caption{Quantitative results of different methods on the dataset of \textit{ImageNet-Test}. The best and second best results are highlighted in \textbf{bold} and \underline{underline}. $*$ indicates that the result was obtained by replicating the method in the paper. Running time is tested on NVIDIA Telsa A100 GPU on the $\times$4 (64$\rightarrow$256) SR tasks. The non-trainable parameters, such as the parameters of VQGAN in LDM, are marked with \textcolor[gray]{0.5}{gray} color for clarity.}
  \label{tab:imagenet_testing}
  \small
  \vspace{-2mm}
  \begin{tabular}{@{}C{4.0cm}@{}|
    @{}C{2.0cm}@{} @{}C{2.0cm}@{} @{}C{2.4cm}@{} @{}C{3.0cm}@{} @{}C{3.5cm}@{}}
      \Xhline{0.8pt}
      \multirow{2}*{Methods} & \multicolumn{5}{c}{Metrics} \\
      \Xcline{2-6}{0.4pt}
          & LPIPS$\downarrow$ & CLIPIQA$\uparrow$ & MUSIQ$\uparrow$ & Runtime (s) & \#Params (M)  \\
          \Xhline{0.4pt}
          ESRGAN~\cite{wang2018esrgan}     & 0.485 & 0.451 & 43.615  & 0.038 & 16.70\\
          RealSR-JPEG~\cite{ji2020real}    & 0.326 & 0.537 & 46.981  & 0.038 & 16.70\\
          BSRGAN~\cite{zhang2021designing} & 0.259 & 0.581 & 54.697  & 0.038 & 16.70\\
          SwinIR~\cite{liang2021swinir}    & 0.238 & 0.564 & 53.790  & 0.107 & 28.01\\
          RealESRGAN~\cite{wang2021real}   & 0.254 & 0.523 & 52.538  & 0.038 & 16.70\\
          DASR~\cite{liang2022efficient}   & 0.250 & 0.536 & 48.337  & 0.022 & 8.06\\
          LDM-15~\cite{rombach2022high}  & 0.269 & 0.512 & 46.419 & 0.408  &{113.60~+~\textcolor[gray]{0.5}{55.32}}  \\
          StableSR-15~\cite{wang2023exploiting}  & 0.262 &\textbf{0.659} &\textbf{59.443} &1.070 &{52.49~+~\textcolor[gray]{0.5}{1422.49}}  \\
          ResShift-15~\cite{yue2024resshift}      & 0.231 & 0.592 & 53.660 & 0.682  &{118.59~+~\textcolor[gray]{0.5}{55.32}} \\
          SinSR-1~\cite{wang2023sinsr}            & \textbf{0.221}  & 0.611  & 53.357 & 0.058 &{118.59~+~\textcolor[gray]{0.5}{55.32}} \\
          SinSR*-1                            & 0.231 & 0.599   & 52.462 & 0.058 &{118.59~+~\textcolor[gray]{0.5}{55.32}}  \\
          \textit{TAD-SR-1}  & \underline{0.227} & \underline{0.652}  & \underline{57.533}  & 0.058 &{118.59~+~\textcolor[gray]{0.5}{57.32}}  \\
     \Xhline{0.8pt}
  \end{tabular} 
\end{table*}

\begin{table*}[t]
  \centering
  \caption{Quantitative results of different methods on two real-world datasets. The best and second best results are highlighted in \textbf{bold} and \underline{underline}.}
  \label{tab:real_testing}
  \small
  \begin{tabular}{@{}C{4.0cm}@{}|
                  @{}C{2.4cm}@{} @{}C{2.5cm}@{}| 
                  @{}C{2.5cm}@{} @{}C{2.5cm}@{} }
      \Xhline{0.8pt}
      \multirow{3}*{Methods} & \multicolumn{4}{c}{Datasets} \\
      \Xcline{2-5}{0.4pt}
          & \multicolumn{2}{c|}{\textit{RealSR}}  & \multicolumn{2}{c}{\textit{RealSet65}} \\
          \Xcline{2-5}{0.4pt}
          & CLIPIQA$\uparrow$ & MUSIQ$\uparrow$   
          & CLIPIQA$\uparrow$ & MUSIQ$\uparrow$  \\
          \Xhline{0.4pt}
          ESRGAN~\cite{wang2018esrgan}     & 0.236 & 29.048   & 0.374 & 42.369  \\
          RealSR-JPEG~\cite{ji2020real}    & 0.362 & 36.076  & 0.528 & 50.539  \\
          BSRGAN~\cite{zhang2021designing} & 0.543 & \underline{63.586} & 0.616 &\underline{65.582} \\
          SwinIR~\cite{liang2021swinir}    & 0.465 & 59.636  & 0.578 & 63.822 \\
          RealESRGAN~\cite{wang2021real}   & 0.490 & 59.678  & 0.600 & 63.220  \\
          DASR~\cite{liang2022efficient}   & 0.363 & 45.825  & 0.497 & 55.708   \\
          LDM-15~\cite{rombach2022high}    & 0.384 & 49.317 & 0.427 & 47.488  \\
          StableSR-15~\cite{wang2023exploiting} & 0.493 & 59.634 & 0.531 & 56.301 \\
          ResShift-15~\cite{yue2024resshift}  & 0.596 & 59.873 & 0.654 & 61.330  \\
          SinSR-1~\cite{wang2023sinsr}       & 0.689  & 61.582  & \underline{0.715} & 62.169  \\
          SinSR*-1                     & \underline{0.691}  & 60.865  & 0.712 & 62.575 \\
          \textit{TAD-SR-1}                           & \textbf{0.741} &\textbf{65.701}  &\textbf{0.734} &\textbf{67.500} \\
      \Xhline{0.4pt}
    \end{tabular} 
\end{table*}

\section{Experiments}
\subsection{Experimental Setup}
\textbf{Training Details.} For a fair comparison, we follow the same experimental setup and backbone design as that in \cite{yue2024resshift,wang2023sinsr}. Specifically, we use the weights of the teacher model (ResShift) to initialize the student model, and then train the model for 30K iterations based on our proposed loss functions. For the real-world image super-resolution task, we set the weighting factor $\lambda_1 = 1$ and $\lambda_2 = 0.02$. For the blind face restoration task, we set $\lambda_1 = 0.1$ and $\lambda = 0.2$.

\textbf{Compared methods.} For real-world SR task, we evalute the effectiveness and effeciency of TAD-SR in comparison to several representative SR models, including BSRGAN \cite{zhang2021designing}, SwinIR \cite{liang2021swinir}, RealESRGAN \cite{wang2021real}, DASR \cite{liang2022efficient}, LDM \cite{rombach2022high}, ResShift \cite{yue2024resshift} and SinSR \cite{wang2023sinsr}. For blind face restoration task, we compare TAD-SR with seven recent BFR methods, including DFDNet \cite{li2020blind}, PSFRGAN \cite{chen2021progressive}, GFPGAN \cite{wang2021towards}, RestoreFormer \cite{wang2022restoreformer}, VQFR \cite{gu2022vqfr}, CodeFormer \cite{zhou2022towards}, and DifFace \cite{yue2022difface}.

\textbf{Metrics.} For real-world super-resolution tasks, we utilize PSNR, SSIM \cite{wang2004image}, and LPIPS \cite{zhang2018unreasonable}, CLIPIQA \cite{wang2022exploring} and MUSIQ \cite{ke2021musiq} to measure the performance of proposed method. For face blind restoration task, besides the above metrics, we also evaluate methods with identity score (IDS), landmark distance (LMD) and FID. On the real-world datasets, it is worth noting non-reference metrics such as CLIPIQA and MUSIQ are more convicing, because they are specifically designed to assess the realism of images in the absense of ground truth. 

\begin{figure*}
\centerline{\includegraphics[width=1.0\linewidth]{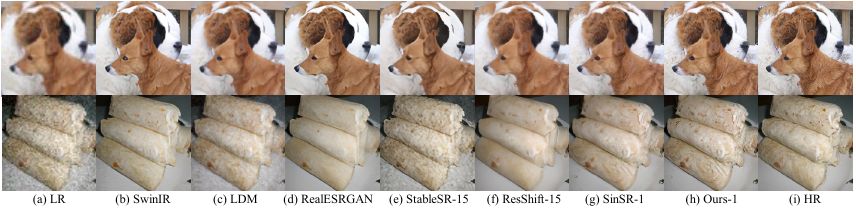}}
\caption{Qualitative comparisons of different methods on two synthetic examples of the \textit{ImageNet-Test} dataset. Please zoom in for a better view.}
\label{fig:imagenet_test}

\end{figure*}

\begin{figure*}
\centerline{\includegraphics[width=1.0\linewidth]{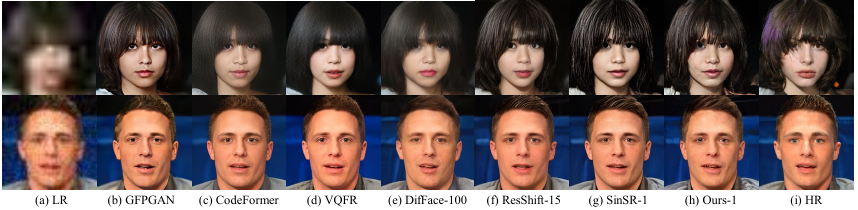}}
\caption{Qualitative comparisons of different methods on two synthetic examples of the \textit{CelebA-Test} dataset. Please zoom in for a better view.}
\label{fig:celeba_test}
\end{figure*}

\textbf{Datasets.} For the real-world image super-resolution task, we train the models on the training set of ImageNet \cite{deng2009imagenet} following the same pipeline with ResShift \cite{yue2024resshift} where the degradation model is adopted from RealESRGAN \cite{wang2021real}. Then, we evaluate our model on one synthetic dataset ImageNet-Test \cite{deng2009imagenet,yue2024resshift} and two real-word datasets RealSR \cite{cai2019toward} and RealSet65 \cite{yue2024resshift}. For the blind face restoration task, We train the models on FFHQ dataset \cite{karras2019style}, and the LQ images are synthesized following a typical degradation model used in \cite{wang2021towards}. One synthetic dataset CelebA-Test \cite{karras2018progressive,yue2024resshift} and three real-world datasets LFW \cite{huang2008labeled}, WebPhoto and WIDER \cite{yang2016wider} are adopted to evaluate the performance of face restoration model.

\subsection{Experimental Results}

\textbf{Evaluation on synthetic datasets.} For the real-world SR task, we conducted a comprehensive comparison between TAD-SR and other state-of-the-art methods on the ImageNet-Test dataset, as summarized in Table~\ref{tab:imagenet_testing} and Fig.~\ref{fig:imagenet_test}. The following conclusions can be drawn: i) TAD-SR significantly outperforms other methods in terms of non-reference metrics, specifically CLIPIQA and MUSIQ, and achieves second-best results in the full-reference metric LPIPS. It demonstrates the effectiveness of the proposed method. ii) The substantial improvement in non-reference metrics indicates that TAD-SR has the ability to generate images with high perceptual quality and realism. iii) Visual results show that TAD-SR produces images with higher clarity and better visual perception. In summary, the proposed TAD-SR effectively distills the teacher model of ResShift into a single step, generating more realistic image results while maintaining comparable fidelity to the teacher model. For blind face restoration, We used CelebA-Test as the testing dataset, and results are summarized in Table~\ref{tab:celeba_testing} and Fig.~\ref{fig:celeba_test}. From the perspective of evaluation metrics, the proposed method achieves state-of-the-art results in terms of FID and comparable results in terms of IDS, LDM, CLIPIQA, and MUSIQ, which demonstrates the effectiveness of TAD-SR on blind face restoration. From the visualization results, the generated faces by the proposed method appear more natural and exhibit richer details. For example, in the images of Fig.~\ref{fig:celeba_test}, the results of TAD-SR show more natural and realistic hair, while the facial clarity is higher, with complete and rich details.

\begin{table*}[t]
	\centering
	\caption{Quantitative results of different methods on the dataset of \textit{CelebA-Test}. The best and second best results are highlighted in \textbf{bold} and \underline{underline}.}
	\label{tab:celeba_testing}
	\small
	\vspace{-2mm}
	\begin{tabular}{@{}C{3.0cm}@{}|
			@{}C{1.5cm}@{} @{}C{1.5cm}@{} @{}C{1.5cm}@{} @{}C{1.6cm}@{} @{}C{1.6cm}@{} @{}C{1.6cm}@{} @{}C{1.6cm}@{}}                                                                                                                         
		\Xhline{0.8pt}
		\multirow{2}*{Methods} & \multicolumn{7}{c}{Metrics} \\
		\Xcline{2-8}{0.4pt}
		& LPIPS$\downarrow$ & IDS$\downarrow$ & LMD$\downarrow$ & FID-F$\downarrow$ & FID-G$\downarrow$ & CLIPIQA$\uparrow$ & MUSIQ$\uparrow$ \\
		\Xhline{0.4pt}
		DFDNet \cite{li2020blind}                           & 0.739 & 86.323 & 20.784 & 93.621 & 76.118 & 0.619 & 51.173  \\
		PSFRGAN  \cite{chen2021progressive}                         & 0.475 & 74.025 & 10.168 & 63.676 & 60.748 & 0.630 & 69.910  \\
		GFPGAN \cite{wang2021towards}                           & 0.416 & 66.820 & 8.886 & 66.308 & 27.698 & 0.671 & \underline{75.388}  \\
		RestoreFormer \cite{wang2022restoreformer}                    & 0.488 & 70.518 & 11.137 & 50.165 & 51.997 & \textbf{0.736} & 71.039  \\
		VQFR \cite{gu2022vqfr}                            & 0.411 & 65.538 & 8.910 & 58.423 & 25.234 & 0.685 & 73.155  \\
		CodeFormer \cite{zhou2022towards}                      & 0.324 & \textbf{59.136} & 5.035 & 62.794 & 26.160 & 0.698 & \textbf{75.900}  \\
		DifFace-100 \cite{yue2022difface}                         & 0.338 & 63.033 & 5.301 & 52.531 & 23.212 & 0.527 & 66.042  \\
		ResShift-4 \cite{yue2024resshift}  & \textbf{0.309} & \underline{59.623} & 5.056 & \underline{50.164} & \underline{17.564} & 0.613 & 73.214  \\
		SinSR*-1 \cite{wang2023sinsr}       & \underline{0.319} & 60.305 & \textbf{4.935} & 55.292 & 21.681 & 0.634 & 74.140  \\
		\textit{TAD-SR-1}                           & 0.341 & 59.897 & \underline{5.050} & \textbf{41.968} & \textbf{16.779} & \underline{0.735} & 75.027  \\
		\Xhline{0.8pt}
	\end{tabular} 
	\vspace{-2mm}
\end{table*}

\begin{table*}[tbp]
	\centering
	\caption{Quantitative results of different methods on three real-world human face datasets. The best and second best results are highlighted in \textbf{bold} and \underline{underline}.}
	\label{tab:real_face_testing}
	\small
	\begin{tabular}{@{}C{3.1cm}@{}|
			@{}C{1.8cm}@{} @{}C{1.8cm}@{}| 
			@{}C{1.8cm}@{} @{}C{1.8cm}@{}|
			@{}C{1.8cm}@{} @{}C{1.8cm}@{} }
		\Xhline{0.8pt}
		\multirow{3}*{Methods} & \multicolumn{6}{c}{Datasets} \\
		\Xcline{2-7}{0.4pt}
		& \multicolumn{2}{c|}{\textit{LFW}}  & \multicolumn{2}{c|}{\textit{WebPhoto}} & \multicolumn{2}{c}{\textit{WIDER}} \\
		\Xcline{2-7}{0.4pt}
		& CLIPIQA$\uparrow$ & MUSIQ$\uparrow$   
		& CLIPIQA$\uparrow$ & MUSIQ$\uparrow$ 
		& CLIPIQA$\uparrow$ & MUSIQ$\uparrow$  \\
		\Xhline{0.4pt}
		DFDNet \cite{li2020blind}                           & 0.716 & 73.109 & 0.654 & 59.024 & 0.625 & 63.210 \\
		PSFRGAN \cite{chen2021progressive}                         & 0.647 & 73.602 & 0.637 & 71.674 & 0.648 & 71.507 \\
		GFPGAN \cite{wang2021towards}                          & 0.687 & \underline{74.836} & 0.651 & \underline{73.369} & 0.663 & \textbf{74.694} \\
		RestoreFormer \cite{wang2022restoreformer}                   & \underline{0.741} & 73.704 & \underline{0.709} & 69.837 & \underline{0.730} & 67.840 \\
		VQFR  \cite{gu2022vqfr}                           & 0.710 & 74.386 & 0.677 & 70.904 & 0.707 & 71.411 \\
		CodeFormer \cite{zhou2022towards}                     & 0.689 & \textbf{75.480} & 0.692 & \textbf{74.004} & 0.699 & 73.404 \\
		DifFace-100 \cite{yue2022difface}                        & 0.593 & 70.362 & 0.555 & 65.379 & 0.561 & 64.970 \\
		ResShift-4 \cite{yue2024resshift}  & 0.626 & 70.643 & 0.621 & 71.007 & 0.629 & 71.084 \\
		SinSR*-1 ~\cite{wang2023sinsr}       & 0.640 & 72.457 & 0.641 & 73.357 & 0.654 & 73.556 \\
		\textit{TAD-SR-1}                           & \textbf{0.768} & 74.085 & \textbf{0.718} & 71.952 & \textbf{0.770} & \underline{73.739} \\
		\Xhline{0.4pt}
	\end{tabular} 
	\vspace{-4mm}
\end{table*}

\textbf{Evaluation on real-world datasets.} For the SR task, in addition to evaluating our method on synthetic datasets, we also evaluated the method in real-world scenarios, and results are reported in Table \ref{tab:real_testing}. As shown in the table, in terms of the CLIPIQA and MUSIQ metrics, the proposed method significantly outperforms other methods with just a single-step inference. Specifically, compared to the ResShift 15-step model serving as our teacher model, these non-reference metrics show a significant improvement after applying TAD-SR. Additionally, visual comparisons are displayed in Fig.~\ref{fig:realworld_appendix}. To ensure a comprehensive evaluation, we try to include diverse scenarios, such as buildings, animals, and letters. It can be observed that the images generated by TAD-SR appear more naturalistic, as evidenced by the distinct brick textures, as well as the fine and natural-looking polar bear fur. For blind face restoration, we evaluated recent methods on three datasets: LFW, Webphoto, and WIDER. The results are presented in Table \ref{tab:real_face_testing}, and significant conclusions can be drawn as follows. On all three datasets, the proposed method achieves the best CLIPIQA, which outperforms the other methods by a large margin. On the WIDER dataset, the proposed method achieves the second-best MUSIQ. All these results inform that in terms of blind face restoration, TAD-SR can generate images with really high perceptual quality. We provide visual comparisons in Fig.~\ref{fig:realworld_face_appendix}. It can be observed that the results obtained by TAD-SR show more realistic hair details, clearer facial contours, and better skin textures.

\subsection{Ablation Study}
The aforementioned experiments have confirmed the effectiveness of our method in image super-resolution tasks. This section is dedicated to presenting ablation studies that aim to further validate the importance of the crucial modules introduced within our framework.

\begin{table}[t]
	\centering
	\caption{Ablation studies of the proposed methods on \textit{RealSR} and \textit{RealSet65} benchmarks. The best results are highlighted in \textbf{bold}.}
	\label{tab:ablation}
	\small
  \begin{tabular}{c|c|c|c}
		\Xhline{0.8pt}
		\multirow{2}*{Methods} & \multirow{2}*{Settings} & \multicolumn{2}{c}{\textit{RealSR}/\textit{RealSet65}} \\
		\Xcline{3-4}{0.4pt}
		 &~ &CLIPIQA$\uparrow$ & MUSIQ$\uparrow$   \\
		\Xhline{0.4pt}
		(a)  & SDS~+~GAN &  0.489/0.528 & 57.290/57.567   \\
		(b)  & SDS~+~t-GAN     &  0.538/0.554 & 60.223/59.627   \\
		(c)  & HSD~+~GAN    &  0.711/0.729 & 63.550/66.904   \\
		Ours  & HSD~+~t-GAN   &  \textbf{0.741/0.734} & \textbf{65.701/67.500}   \\
		\Xhline{0.4pt}
	\end{tabular} 
	\vspace{-4mm}
\end{table}

\textbf{High-frequency enhanced score distillation.} We first investigate the importance of high-frequency enhanced score distillation. Recall that in Section~\ref{sub:TAD-SR}, we analyzed how high-frequency enhanced score distillation can provide a meaningful guidance for optimizing student model compared to score distillation sampling (SDS). Here, we further validate its effectiveness on real-world benchmarks. As shown in Table \ref{tab:ablation}, compared with SDS, our proposed high-frequency enhanced score distillation can significantly improved the CILIPIQA and MUSIQ scores on both the RealSR and RealSet65 datasets. This indicates that our proposed high-frequency enhanced score distillation can indeed improve the quality of image generation and is superior to SDS.

\textbf{Time-aware discriminator.} It has been proven that introducing generative adversarial training in latent space is easier to optimize and more cost-effective compared to pixel space \cite{sauer2024fast}. Now, we demonstrate the importance of introducing time injection into the discriminator. Intuitively, when the discriminator does not have time injection, it needs to distinguish the distribution between real data and generated data under different noise disturbances, which is undoubtedly extremely challenging. Adding time injection to the discriminator is equivalent to providing additional information related to the level of noise disturbance, which should improve the performance of the discriminator and provide more effective supervision for the generator. We further validated the above analysis through experiments. As shown in Table \ref{tab:ablation}, performance improves with the replacement of the standard discriminator by our proposed time-aware discriminator, regardless of the score distillation technique used.

\section{Conclusion}
In this paper, we propose a time-aware distillation method that accelerates diffusion-based super-resolution models to a single inference step. We introduce a high-frequency enhanced score distillation technique that optimizes the generator by calculating the score difference between the outputs of the teacher and student models following noise perturbation, thereby enhancing the high-frequency details in the student model's output. To elevate the student model's performance ceiling, we incorporate generative adversarial learning into the diffusion model framework. Specifically, we design a time-aware discriminator that distinguishes between generated and real data in latent space, providing more efficient and effective supervision for the student model. Extensive experiments demonstrate that our method can achieve performance on par with or surpassing that of the teacher model in a single inference step.

\bibliography{reference}
\bibliographystyle{IEEEtran}

\clearpage
\onecolumn
\appendix
\subsection{Implementation Details}

\subsubsection{Mathematical Details} \label{subsec:math_supp}
\begin{itemize}[topsep=0pt,parsep=0pt,leftmargin=18pt]
    \item \textbf{Derivation of Eq.~\eqref{eq:HSD_V2}}:
    According to the transition distribution of Eq.~\eqref{eq:forward} of our manuscript, the predicted noise $\epsilon_{\phi}$ can be expressed via the following reparameterization trick:
    \begin{equation}
        \epsilon_{\phi} = \frac{z_t - \left( \hat{z}_0 + \eta_t\left( z_y - \hat{z}_0 \right) \right)}{\sqrt{\eta_t \kappa}},
        \label{eq:reparametrization_eps}
    \end{equation}
    where $\hat{z}_0 = F_{\phi}\left(z_t,t,y\right)$.
    According to the Eq.~\eqref{eq:reparametrization_eps}, we can rewrite Eq.~\eqref{eq:HSD_V1} as follows:

    \begin{equation}
      \mathcal{L}_{hsd} = \mathbb{E}_{z_{t^{'}}^{tch},z_{t^{'}}^{stu},y} \left[ \frac{\omega \left(\left(z_{t^{'}}^{stu}-z_{t^{'}}^{tch}\right)+\left(1-\eta_{t^{'}}\right)\left(F_{\phi}\left(z_{t^{'}}^{tch},t^{'},y\right)-F_{\phi}\left(z_{t^{'}}^{stu},t^{'},y\right)\right)\right)}{\sqrt{\eta_{t^{'}}}\kappa}\right].
      \label{eq:hsd_inter}
  \end{equation}
  Since the noise injected into the output image of the student model and the output image of the teacher model is the same, we have: $z_{t^{'}}^{stu}-z_{t^{'}}^{tch}= \left(1-\eta_{t^{'}}\right)\left(z_{0}^{stu}-z_{0}^{tch}\right)$. Then Eq.~\eqref{eq:hsd_inter} can be written as:

  \begin{align}
    \mathcal{L}_{hsd} &= \mathbb{E}_{z_{t^{'}}^{tch},z_{t^{'}}^{stu},y} \left[ \frac{\omega \left(1-\eta_{t^{'}}\right)\left(z_{0}^{stu}-z_{0}^{tch}+F_{\phi}\left(z_{t^{'}}^{tch},t^{'},y\right)-F_{\phi}\left(z_{t^{'}}^{stu},t^{'},y\right)\right)}{\sqrt{\eta_{t^{'}}}\kappa}\right] \notag \\
      &= \mathbb{E}_{z_{t^{'}}^{tch},z_{t^{'}}^{stu},y} \left[\omega_2 \left(z_{0}^{stu}-z_{0}^{tch}+F_{\phi}\left(z_{t^{'}}^{tch},t^{'},y\right)-F_{\phi}\left(z_{t^{'}}^{stu},t^{'},y\right)\right)\right],
    \label{eq:hsd_final}
\end{align}
where $\omega_2 = \frac{\omega \left(1-\eta_{t^{'}}\right)}{\sqrt{\eta_{t^{'}}}}$
\end{itemize}

\subsubsection{TAD-SR Training Procedure}
For a comprehensive understanding, we provide a detailed description of our TAD-SR training procedure in Algorithm~\ref{alg:distillation}.
\begin{algorithm}
    \caption{\label{alg:distillation}TAD-SR Training Procedure}
    \KwIn{Pretrained diffusion model $F_{\phi}$,
    paired dataset $\mathcal{D}=\{x, y\}$, Time steps $T$
    }
    \KwOut{Trained generator $f_{\theta}$ and disriminator $D_{\psi}$.}
    
    \tcp{Initialize generator from pretrained model}
    $f_{\theta} \leftarrow \text{copyWeights}(F_{\Phi}),$

    \While{train}{
        \tcp{Generated images}
        Sample $\epsilon \sim\mathcal{N}(0, \mathbf{I}),~(x, y) \sim\mathcal{D}$
        
        $z_{T} \leftarrow \text{Forward process}(T,y,x,\epsilon)$ \tcp{Eq~\eqref{eq:forward}}

        $z_{0}^{stu} \leftarrow f_{\theta}(z_T,y,T)$  \tcp{One-step}
        
        $z_{0}^{tch} \leftarrow F_{\phi}(z_T,y,T)$  \tcp{Multi-step}

        \text{~}
        
        \tcp{Update discriminator model}
        Sample time step $t\sim\mathcal{U}(0,T)$

        $z_{t}^{stu} \leftarrow \text{Forward process}(t,y,z_{0}^{stu},\epsilon)$ \tcp{Eq~\eqref{eq:forward}}

        $z_{t}  \leftarrow \text{Forward process}(t,y,x,\epsilon)$ \tcp{Eq~\eqref{eq:forward}}

        $\mathcal{L}_\text{adv}^{D_{\psi}} \leftarrow \text{Adversarial loss}(z_{t}^{stu}, z_{t},y,t) $ \tcp{Eq~\eqref{eq:adv_D}}

        $D_{\psi} \leftarrow \text{update}(D_{\psi}, \mathcal{L}_\text{adv}^{D_{\psi}})$

        \text{~}
        
        \tcp{Update generator}
        Sample $\epsilon^{'} \sim\mathcal{N}(0, \mathbf{I}),t^{'}\sim\mathcal{U}(0,T/5)$

        $z_{t^{'}}^{stu} \leftarrow \text{Forward process}(t^{'},y,z_{0}^{stu},\epsilon^{'})$ \tcp{Eq~\eqref{eq:forward}}

        $z_{t^{'}}^{tch}  \leftarrow \text{Forward process}(t^{'},y,z_{0}^{tch},\epsilon^{'})$ \tcp{Eq~\eqref{eq:forward}}

        $\mathcal{L}_\text{distill} \leftarrow \text{Vanilla distillation Loss}(z_{0}^{stu}, z_{0}^{tch})$ \tcp{Eq~\eqref{eq:HSD_V2}}
        
        $\mathcal{L}_\text{hsd} \leftarrow \text{HSD loss}(z_{t^{'}}^{stu}, z_{t^{'}}^{tch},y,t^{'})$ \tcp{Eq~\eqref{eq:distill}}
        
        $\mathcal{L}_\text{adv}^{f_{\theta}} \leftarrow \text{Adversarial loss}(z_{t}^{stu},y,t)$ \tcp{Eq~\eqref{eq:adv_G}}

        $\mathcal{L}_{f_{\theta}} \leftarrow \mathcal{L}_\text{distill} + \lambda_1\mathcal{L}_\text{hsd} + \lambda_2\mathcal{L}_\text{adv}^{f_{\theta}}$

        $f_{\theta} \leftarrow \text{update}(f_{\theta}, \mathcal{L}_{f_{\theta}})$
        
    }
\end{algorithm}

\subsection{Experimental Results on SD-based SR method}
\subsubsection{Experimental Setup} \label{subsec:exp_bfr_setup} \
In addition to distilling the super-resolution model trained from scratch, we also apply our proposed TAD-SR to distill the SOTA SD-based super-resolution model to further validate its effectiveness.

\noindent\textbf{Training Datasets}.
We adopt DIV2K \cite{agustsson2017ntire}, Flickr2K \cite{timofte2017ntire}, first 20K images from LSDIR \cite{li2023lsdir} and first 10K face images from FFHQ \cite{karras2019style} for training. The degradation pipeline of Real-ESRGAN \cite{wang2021real} is used to synthesize LR-HR training pairs.

\noindent\textbf{Testing Datasets}. 
We evaluate TAD-SR on three real-world datasets: DRealSR \cite{wei2020component}, RealSR \cite{cai2019toward}, and RealLR200 \cite{wu2024seesr}, as well as one synthetic dataset, DIV2K-val\cite{agustsson2017ntire}. The method for acquiring HR-LR image pairs in the DIV2K dataset follows the procedure detailed in \cite{wang2023exploiting}, and except RealLR200, all datasets are cropped to 512×512 patches.

\noindent\textbf{Compared Methods.} 
We compare our SeeSR with several state-of-the-art Real-ISR methods, which can be categorized into two groups. The first group consists of GAN-based methods, including BSRGAN \cite{zhang2021designing}, Real-ESRGAN \cite{karras2019style}, LDL \cite{liang2022details}, FeMaSR \cite{chen2022real}. The second group consists of recent diffusion-based methods, including StableSR \cite{wang2023exploiting}, ResShift \cite{yue2024resshift}, PASD \cite{yang2023pixel}, SeeSR \cite{wu2024seesr}, SinSR \cite{wang2023sinsr} and AddSR \cite{xie2024addsr}. 

\noindent\textbf{Evaluation Metrics.} 
We employ non-reference metrics (e.g., MANIQA \cite{yang2022maniqa}, MUSIQ \cite{ke2021musiq}, CLIPIQA \cite{wang2022exploring}) and reference metrics (e.g., LPIPS \cite{zhang2018learning}, PSNR, SSIM \cite{wang2004image}) to comprehensively evaluate our TAD-SR. Note that in real-world super-resolution tasks, the non-reference metrics is more aligned with human perception and better reflects the subjective quality of images.

\begin{table*}
  \centering
  \caption{Quantitative comparison with state of the arts on RealSR dataset and RealLR dataset. The best and second best results are highlighted in \textbf{bold} and \underline{underline}. Note that since the RealLR200 dataset lacks real high-resolution images, we only computed non-reference metrics.}
  \label{tab:RealSR_testing}
  \begin{tabular}{c|cccccc|ccc}
      \Xhline{0.8pt}
      \multirow{2}*{Methods} & \multicolumn{6}{c}{RealSR} & \multicolumn{3}{c}{RealLR} \\
      \Xcline{2-10}{0.4pt}
          &PSNR$\uparrow$ & SSIM $\uparrow$ & LPIPS $\downarrow$ & CLIPIQA$\uparrow$ & MUSIQ$\uparrow$   & MANIQA$\uparrow$
          & CLIPIQA$\uparrow$ & MUSIQ$\uparrow$  & MANIQA$\uparrow$ \\
          \Xhline{0.4pt}
          BSRGAN~\cite{zhang2021designing}     & \underline{26.49} & \textbf{0.767}   & \textbf{0.267} & 0.512 & 63.28 & 0.376 & 0.570 & 64.87 & 0.369   \\
          RealESRGAN~\cite{wang2021real}   & 25.78 & 0.762  & \underline{0.273} & 0.449 & 60.36 & 0.373 & 0.542 & 62.93 & 0.366  \\
          LDL~\cite{liang2022details} & 25.09 & \underline{0.764} & 0.277 & 0.430 & 58.04 & 0.342 & 0.509 & 60.95 & 0.327 \\
          FeMaSR~\cite{chen2022real}  & 25.17 & 0.736  & 0.294 & 0.541 & 59.06 & 0.361 & 0.655 & 64.24  & 0.410  \\
          \Xhline{0.4pt}
          StableSR-200~\cite{wang2023exploiting} & 25.63 & 0.748 & 0.302 & 0.528 & 61.11 & 0.366 & 0.592 & 62.89 & 0.367 \\
          ResShift-15~\cite{yue2024resshift} & 26.34 & 0.735  & 0.346 & 0.542 & 56.06 & 0.375 & 0.647 & 60.25 & 0.418  \\
          PASD-20~\cite{yang2023pixel} & \textbf{26.67} & 0.758 & 0.344 & 0.519 & 62.92 & 0.404 & 0.620 & 66.35 & 0.419 \\\
          SeeSR-50 (teacher)~\cite{wu2024seesr} & 25.24 & 0.720 & 0.301 & \underline{0.670} & \textbf{69.82} & \textbf{0.540} & 0.662 & \underline{68.63} & \textbf{0.491} \\
          SinSR-1~\cite{wang2023sinsr}   & 26.16 & 0.739 & 0.308 & 0.630  & 60.96  & 0.399  & \textbf{0.697} & 63.85 & 0.445  \\
          AddSR-1~\cite{xie2024addsr}    & 23.12 & 0.655 & 0.309 & 0.552  & 67.14 & 0.488       & 0.585  & 66.86  & 0.418  \\
          TAD-SR-1 (Ours)   & 24.50 & 0.710 & 0.304 & \textbf{0.676} & \underline{69.02} & \underline{0.526} & \underline{0.674} & \textbf{69.48} & \underline{0.482}  \\
    \Xhline{0.4pt}
  \end{tabular} 
\end{table*}
\begin{table*}
  \centering
  \caption{Quantitative comparison with state of the arts on DRealSR dataset. The best and second best results are highlighted in \textbf{bold} and \underline{underline}.}
  \label{tab:DRealSR_testing}
  \begin{tabular}{c|cccccc}
      \Xhline{0.8pt}
      \multirow{2}*{Methods} & \multicolumn{6}{c}{DRealSR} \\
      \Xcline{2-7}{0.4pt}
          &PSNR$\uparrow$ & SSIM $\uparrow$ & LPIPS $\downarrow$ & CLIPIQA$\uparrow$ & MUSIQ$\uparrow$   & MANIQA$\uparrow$
           \\
          \Xhline{0.4pt}
          BSRGAN~\cite{zhang2021designing}     & 28.68 & 0.802   & 0.288 & 0.509 & 57.17 & 0.343   \\
          RealESRGAN~\cite{wang2021real}   & 28.57 & \underline{0.804}  & \underline{0.285} & 0.451 & 54.27 & 0.343   \\
          LDL~\cite{liang2022details} & 27.41 & \textbf{0.807} & \textbf{0.282} & 0.441 & 52.38 & 0.324  \\
          FeMaSR~\cite{chen2022real}  & 26.83 & 0.755  & 0.317 & 0.564 & 53.70 & 0.318  \\
          \Xhline{0.4pt}
          StableSR-200~\cite{wang2023exploiting} & \textbf{29.14} & \underline{0.804} & 0.332 & 0.510 & 52.28 & 0.322  \\
          ResShift-15~\cite{yue2024resshift} & 28.27 & 0.754  & 0.401 & 0.529 & 50.14 & 0.328  \\
          PASD-20~\cite{yang2023pixel} & \underline{29.06} & 0.791 & 0.393 & 0.538 & 55.33 & 0.387 \\
          SeeSR-50 (teacher)~\cite{wu2024seesr} & 28.09 & 0.766 & 0.319 & \textbf{0.691} & \textbf{65.08} & \textbf{0.513}  \\
          SinSR-1~\cite{wang2023sinsr}   & 28.32 & 0.747 & 0.372 & 0.642  & 55.36  & 0.388  \\
          AddSR-1~\cite{xie2024addsr}    & 26.70 & 0.738 & 0.320 & 0.593  & 62.13 & 0.458   \\
          TAD-SR-1 (Ours)   & 27.62 & 0.767 & 0.315 & \underline{0.678} & \underline{63.43} & \underline{0.475} \\
    \Xhline{0.4pt}
  \end{tabular} 
\end{table*}

\begin{table*}
  \centering
  \caption{Quantitative comparison with state of the arts on DIV2k-val dataset. The best and second best results are highlighted in \textbf{bold} and \underline{underline}.}
  \label{tab:div2k_testing}
  \begin{tabular}{c|cccccc}
      \Xhline{0.8pt}
      \multirow{2}*{Methods} & \multicolumn{6}{c}{DIV2K-val} \\
      \Xcline{2-7}{0.4pt}
          &PSNR$\uparrow$ & SSIM $\uparrow$ & LPIPS $\downarrow$ & CLIPIQA$\uparrow$ & MUSIQ$\uparrow$   & MANIQA$\uparrow$
           \\
          \Xhline{0.4pt}
          BSRGAN~\cite{zhang2021designing}     & \underline{24.58} & 0.627   & 0.335 & 0.524 & 61.19 & 0.356   \\
          RealESRGAN~\cite{wang2021real}   & 24.29 & \textbf{0.637}  & \textbf{0.311} & 0.527 & 61.06 & 0.382   \\
          LDL~\cite{liang2022details} & 23.83 & \underline{0.634} & 0.326 & 0.518 & 60.04 & 0.375  \\
          FeMaSR~\cite{chen2022real}  & 23.06 & 0.589  & 0.346 & 0.599 & 60.82 & 0.346  \\
          \Xhline{0.4pt}
          StableSR-200~\cite{wang2023exploiting} & 23.29 & 0.573 & \underline{0.312} & \underline{0.676} & 65.83 & 0.422  \\
          ResShift-15~\cite{yue2024resshift} & \textbf{24.72} & 0.623  & 0.340 & 0.594 & 60.89 & 0.399  \\
          PASD-20~\cite{yang2023pixel} & 24.51 & 0.627 & 0.392 & 0.551 & 59.99 & 0.399 \\
          SeeSR-50 (teacher)~\cite{wu2024seesr} & 23.68 & 0.604 & 0.319 & \textbf{0.693} & \textbf{68.68} & \textbf{0.504}  \\
          SinSR-1~\cite{wang2023sinsr}   & 24.41 & 0.602 & 0.324 & 0.648  & 62.80  & 0.424  \\
          AddSR-1~\cite{xie2024addsr}    & 23.26 & 0.590 & 0.362 & 0.573  & 63.69 & 0.405   \\
          TAD-SR-1 (Ours)   & 23.54 & 0.63 & \textbf{0.311} & 0.664 & \underline{67.01} & \underline{0.470} \\
    \Xhline{0.4pt}
  \end{tabular} 
\end{table*}

\begin{figure}
	\centerline{\includegraphics[width=0.9\linewidth]{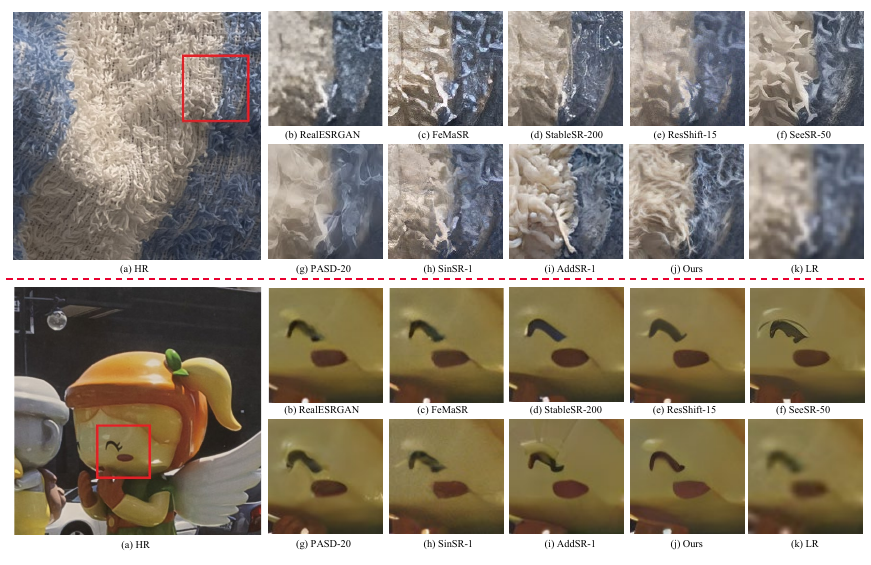}}
	\caption{Visual comparison on real-world LR images. Note that SeeSR is the teacher model.}
        \label{fig:seesr_results}
	
\end{figure}

\subsubsection{Evaluation results}
We first show the quantitative comparison on the four synthetic and real-world datasets in Tables~\ref{tab:RealSR_testing}, ~\ref{tab:DRealSR_testing} and~\ref{tab:div2k_testing}. The observations from the table are as follows: (1) The GAN-based method shows advantages over diffusion-based methods in full-reference metrics, yet it significantly lags behind diffusion-based methods in non-reference metrics. (2) Our method achieves performance comparable to SeeSR using only single-step sampling. (3) Our method attains optimal or near-optimal performance on non-reference metrics. Additionally, the visualization results demonstrate that our method not only enhances image details with greater clarity (as illustrated in the second row of Fig.~\ref{fig:seesr_results}) but also preserves the similarity to the original image as much as possible (as shown in the fourth row of Fig.~\ref{fig:seesr_results}). Overall, compared to other DM-based methods, our TAD-SR achieves comparable or even superior performance through single-step sampling alone.

\subsection{More visualization results}

\begin{figure}
	\centerline{\includegraphics[width=0.9\linewidth]{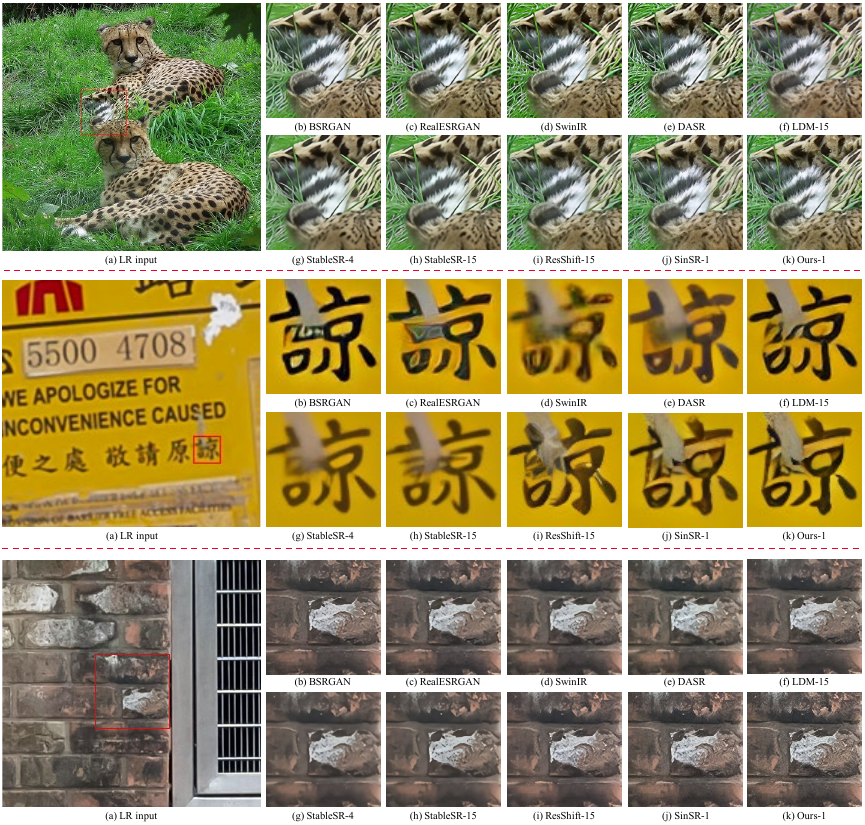}}
	\caption{Qualitative comparisons of different methods on three real-world examples of the \textit{RealSR} and \textit{RealSet65} dataset. Please zoom in for a better view.}
        \label{fig:realworld_appendix}
	
\end{figure}

\begin{figure}
	\centerline{\includegraphics[width=0.85\linewidth]{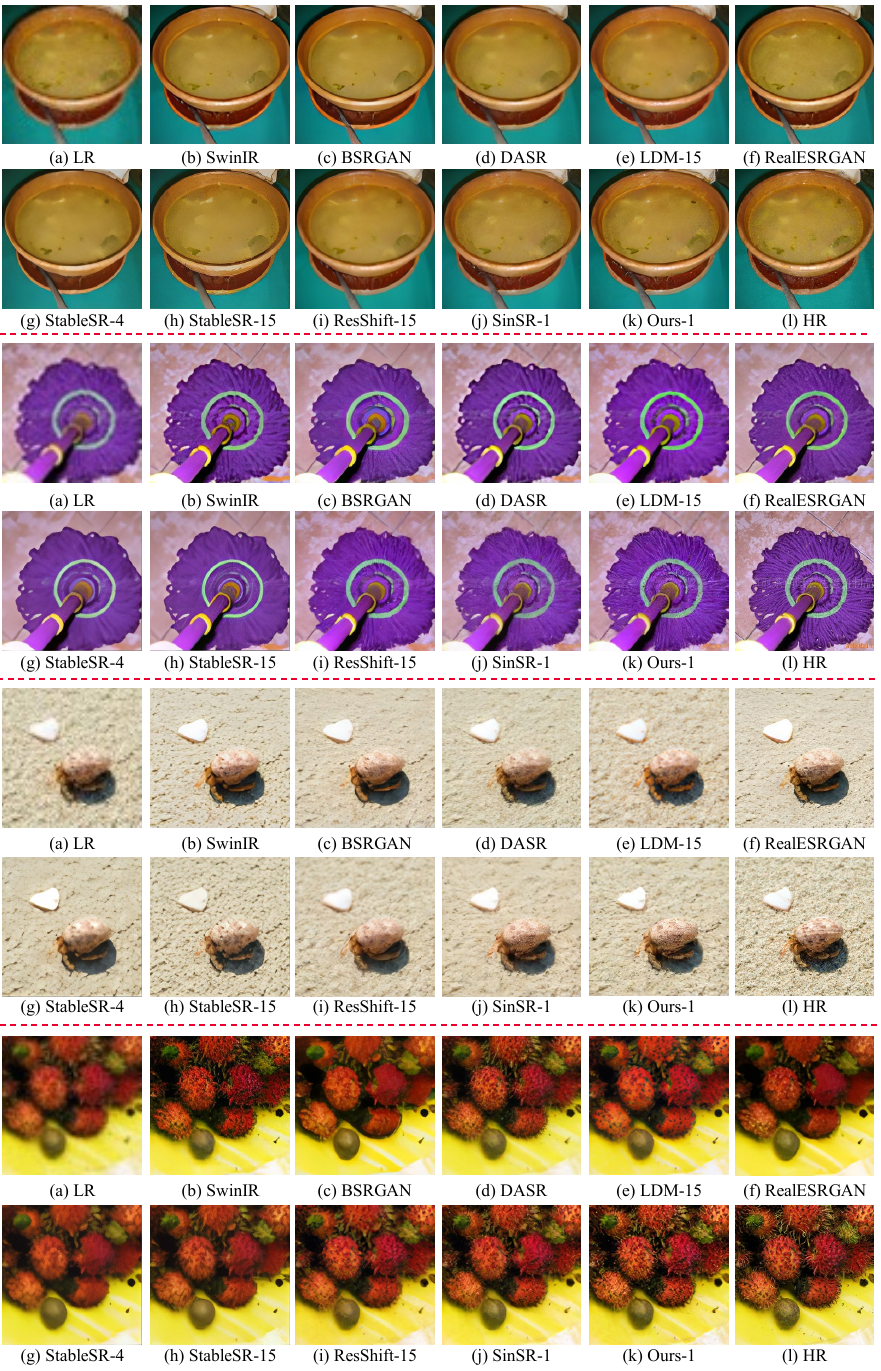}}
	\caption{Qualitative comparisons of different methods on four synthetic examples of the \textit{ImageNet-Test} dataset. }
        \label{fig:imagenet_appendix}
\end{figure}

\begin{figure}
	\centerline{\includegraphics[width=0.85\linewidth]{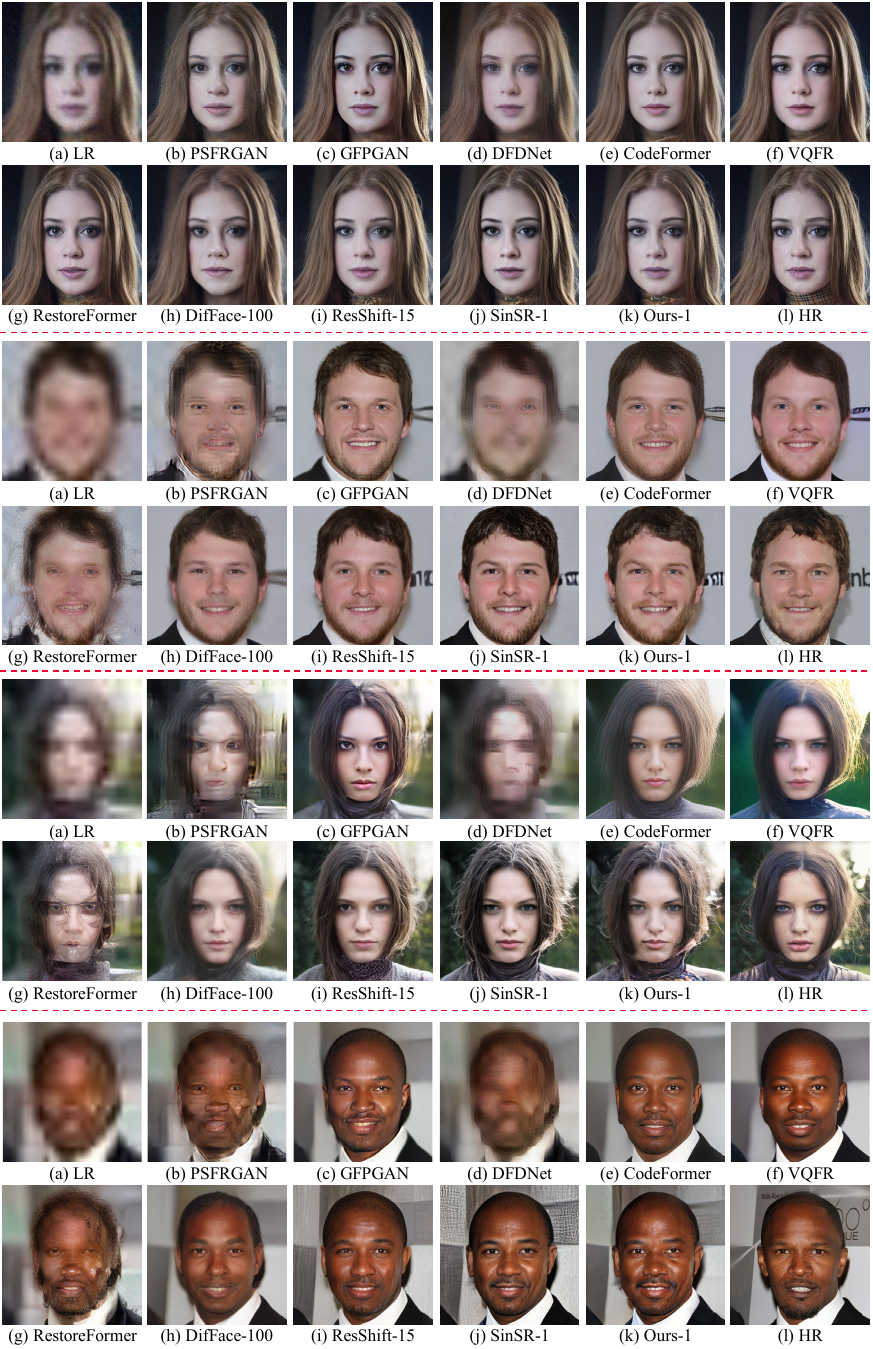}}
	\caption{Qualitative comparisons of different methods on four synthetic examples of the \textit{CelebA-Test} dataset.}
        \label{fig:celebA_appendix}
	\vspace{-4mm}
\end{figure}

\begin{figure}
	\centerline{\includegraphics[width=0.85\linewidth]{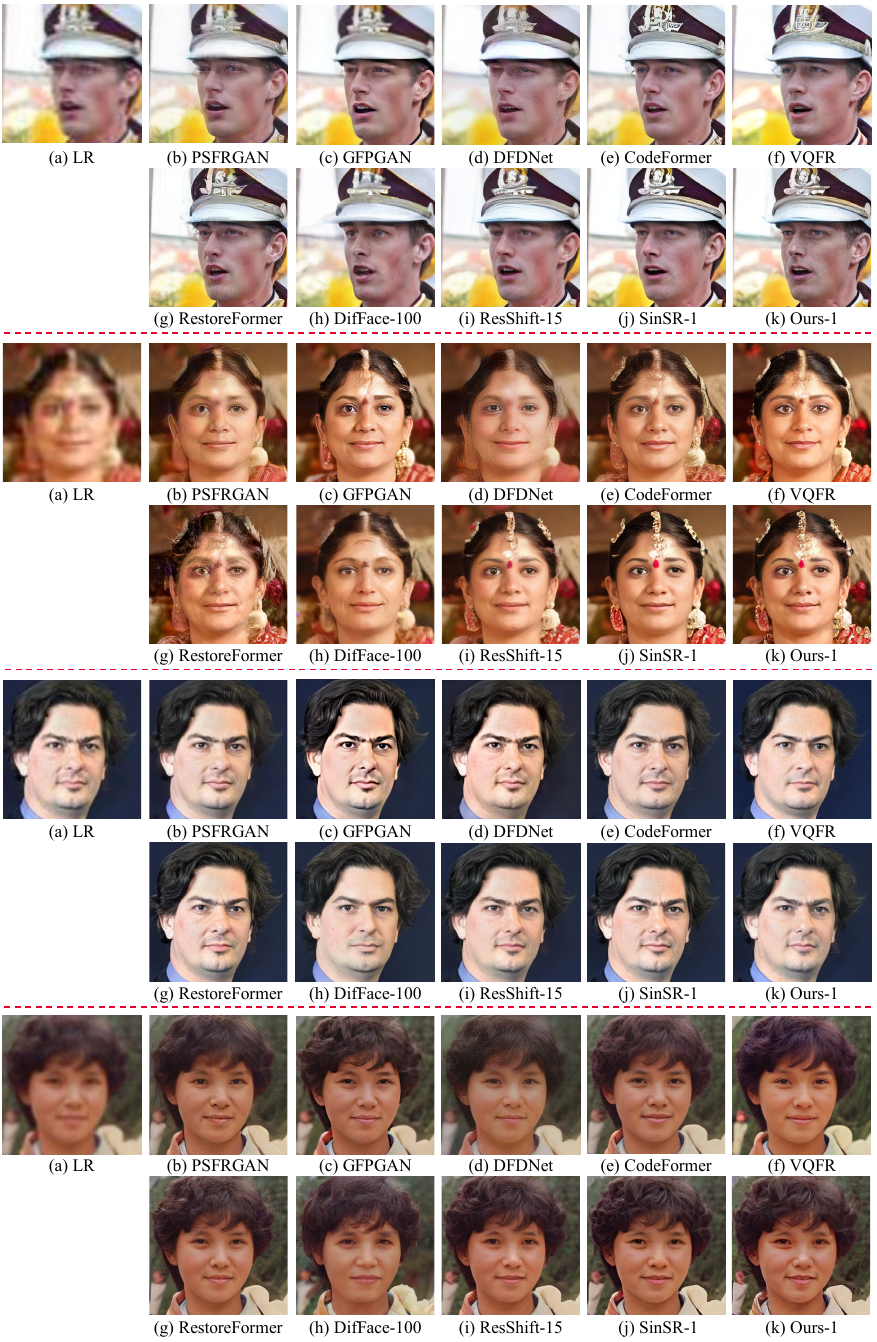}}
	\caption{Qualitative comparisons of different methods on four real-world examples of the \textit{LFW}, \textit{WebPhoto} and \textit{WIDER} dataset.}
        \label{fig:realworld_face_appendix}
	\vspace{-4mm}
\end{figure}

\end{document}